\documentclass[runningheads]{llncs}
\usepackage{graphicx}
\usepackage{subfig}
\usepackage{graphicx}
\usepackage{amsmath}
\usepackage[numbers]{natbib}
\usepackage{amsfonts} 
\begin{document}
\title{The Blame Problem in Evaluating Local Explanations and How to Tackle It}
\titlerunning{The Blame Problem}
%
\author{Amir Hossein Akhavan Rahnama\inst{1}\orcidID{0000-0002-6846-5707}}
\authorrunning{A. H. A. Rahnama}
%
\institute{KTH Royal Institute of Technology\\
\email{amiakh@kth.se}}
\maketitle              
\begin{abstract}
The number of local model-agnostic explanation techniques proposed has grown rapidly recently. One main reason is that the bar for developing new explainability techniques is low due to the lack of optimal evaluation measures. Without rigorous measures, it is hard to have concrete evidence of whether the new explanation techniques can significantly outperform their predecessors. Our study proposes a new taxonomy for evaluating local explanations: robustness, evaluation using ground truth from synthetic datasets and interpretable models, model randomization, and human-grounded evaluation. Using this proposed taxonomy, we highlight that all categories of evaluation methods, except those based on the ground truth from interpretable models, suffer from a problem we call the "blame problem." In our study, we argue that this category of evaluation measure is a more reasonable method for evaluating local model-agnostic explanations. However, we show that even this category of evaluation measures has further limitations. The evaluation of local explanations remains an open research problem. 

\end{abstract}

\section{Introduction}
\label{sec:introduction}
One of the most popular areas within explainable AI is the study of local model-agnostic explanation techniques\footnote{For brevity, we refer to them as local explanations in our study.} \cite{guidotti2021principles}. Local explanations, originally called explanations for individual instances \cite{strumbelj2010efficient}, differ from global explanations. Global explanations are the information intrinsically available in the interpretable models, such as the weights of linear models or the feature importance scores in tree models \cite{freitas2014comprehensible}. Moreover, they provide information about the internal logic of their models at the dataset level, i.e., for all data instances. On the other hand, local explanations are information about the prediction of an individual instance \cite{ribeiro2016model}. One of the main arguments for the need for local explanations is that obtaining a global explanation of complex black-box models for all instances might be hard \cite{ribeiro2016model, ribeiro2016should}. 

There are numerous different ways to represent local explanations. However, feature attribution is the most common representation in the literature of explainable AI \cite{guidotti2021principles}. The feature attribution explanation technique allocates importance scores to each feature, showing their contribution to the predicted output of a black-box model\footnote{See Section \ref{sec:background} for a formal definition of these techniques.}. 

The need for rigorous evaluation of local explanations has been amplified after several studies have shown that local explanations can fail. For example, in \cite{rudin2018please}, the author argues that we should not use local explanation techniques in high-stake decision-making domains by showing numerous failure cases of these techniques. In \cite{molnar2022general}, the authors show that local explanation techniques can fail to consider feature interaction in their output explanations. Meanwhile, the number of proposed local explanation techniques is growing rapidly. In only one study, \cite{covert2021explaining}, the authors have listed 29 local explanation techniques. 

Even though we share the same concern with the authors of \cite{molnar2022general, rudin2018please}, we believe that the real problem that hinders the adaptation of these techniques into high-stake domains is \textit{the lack of optimal measures to evaluate them}. In the absence of strict and rigorous measures, there has been a surge in studies that propose new explainability techniques, yet it is unclear whether the newly introduced explanation techniques can significantly improve upon their predecessors and, if so, in what type of tasks or problems. In \cite{leavitt2020towards}, the authors share their concerns about the poor evaluation of local \textit{model-based} explanation techniques of neural network models. In their words, "interpretability research suffers from an over-reliance on intuition-based approaches that risk-and in some cases have caused illusory progress and misleading conclusions.". 

In an ideal world, each instance will have a local ground truth importance score\footnote{For brevity, we refer to local ground truth importance scores as ground truth. Note that these ground truth vectors differ from the common ground truth in machine learning, which are discrete class labels for the data points.}. These include information about the importance of each feature in the explained instance to the black box's predicted output. We expect this local ground truth to be unique for data instances with substantially different feature values and predicted output. However, we are faced with a contradiction: if we can extract such information from a black box model, why do we call that model a black box, and hence, what is the need for local explanations? Therefore, we can naturally assume a limit to how fine-grained and accurate these ground truth importance scores can be. 

Faced with this problem, researchers in explainable machine learning have introduced and extensively used alternative measures to circumvent this challenging problem. However, we have noticed that these measures' assumptions and limitations are mostly stated implicitly. As a result, researchers can draw misleading conclusions about the accuracy of local explanations when using these evaluation measures. Because of this, in our study, we propose to categorize these evaluation measures based on the assumptions they are based on:

\begin{itemize}
    \item \textbf{Robustness Measures} are based on the \textit{assumption} that nullifying important (unimportant) features should cause large (insignificant) changes to the predicted output of a black-box model for the explained instance. Another sub-category of these measures is based on the assumption that adding a small variation (noise) to the explained instance needs to cause minimal changes to their explanation.
    \item \textbf{Ground Truth from Synthetic Data} is based on the \textit{assumption}  that local explanations must provide feature importance scores similar to the prior importance scores generated by the synthetic data generators. 
    \item \textbf{Ground Truth from Interpretable Models} is based on the \textit{assumption}  that local explanations must be able to allocate similar importance scores to the local ground truth importance scores obtained from simpler and more interpretable models.
    \item \textbf{Model Randomization} evaluation measures are based on the assumption that local explanations of a randomized (contaminated) model must be substantially different from those obtained for the original black-box model.
    \item \textbf{Human-grounded Evaluation} is based on the \textit{assumption}  that if human subjects need to be able to replay the model prediction of a black-box model using local explanations or the content of local explanations need to be similar to the human reasoning process for local explanation to be accurate. 
\end{itemize}

The main contribution of our study is to highlight the different ways we evaluate local explanations. This is in contrast to some studies that have primarily focused on one class of evaluation measures, namely Robustness measures \cite{montavon2018methods}. Moreover, we show all categories of evaluation measures suffer from a range of implicit limitations, the most influential of which is a "blame problem.". In our definition, \textit{the blame problem is when we are unsure whether we should allocate the poor performance of local explanations to themselves or the black-box model}. What becomes straightforward through our proposed systematic categorization is the realization that the "blame problem" is a recurring problem in all the above categories of evaluation methods, except when we evaluate local explanations via extracting ground truth using interpretable models. However, even in this category, we face further limitations. To our knowledge, no study has highlighted the blame problem or systematically investigated these different categories of evaluation measures. Moreover, the relationship between different evaluation measures across these categories is poorly studied. We provide a synthetic example of such investigations.

The rest of the paper is organized as follows. Section \ref{sec:related_work} discusses the related work. In Section \ref{sec:background}, we provide the formal definition of local model-agnostic explanations.  In Section \ref{sec:history}, we briefly discuss how global explanations of black-box models are evaluated, as these methods precede local explanations. In Section, \ref{sec:evaluation}, we present our proposed categories of evaluation measures along with their strengths and limitations. We conclude our study and provide directions for future studies in Section \ref{sec:conclusion}.

\section{Related Work}
\label{sec:related_work}

Several well-written surveys on the local model-agnostic explanation techniques exist in the literature of explainability \cite{guidotti2021principles, molnar2020interpretable}. In \citet{molnar2022general}, the authors provide general pitfalls of local explanation models. However, the study focuses on the limitations of the explanation techniques and not the evaluation measures. 

In \cite{doshi2017towards}, the evaluation methods of local explanations are categorized into Human-grounded, Application-grounded, and Functionally-grounded evaluation. Human-grounded evaluations use expert human subjects to replay the black-box prediction, whereas functionally grounded evaluation uses systematic proxy measures to evaluate local explanations. Application-grounded Evaluation is similar to Human-grounded assessment. However, the experiments involve lay humans, not experts. The first four categories of evaluation we discuss in our study fall into the Functionally-grounded evaluation measure.

In the rest of this section, we provide an overview of the studies that have provided critical overviews or surveys of the evaluation measures. The discussion of related work for each category of evaluation measures will be presented in their respective subsections of Section \ref{sec:evaluation}.  

In \cite{zhou2021evaluating, nauta2023anecdotal}, the authors have focused on providing systematic surveys of the evaluation measures for local explanations. In these surveys, the authors aim to provide a reference for the type of evaluation measures used in the literature on explainability. We consider these studies reliable references for knowing what measures were used to evaluate explanations in the literature. However, our study aims to highlight implicit assumptions and limitations behind these evaluation measures and their respective limitations. 

In \cite{hancox2020robustness}, the author argues that robustness analysis of local explanations is useful for obtaining explanations that can generalize to real-world problems. In \cite{hedstrom2023quantus}, the authors propose an evaluation toolkit for evaluating the local model-based explanations of neural networks. In \cite{agarwal2022openxai}, the authors propose an open benchmark to evaluate local explanations. Most of the measures included in the two aforementioned studies are based on the robustness analysis (See Section \ref{sec:evaluation:robustness} for more details on robustness analysis). 

In \cite{leavitt2020towards}, the authors state four criticisms of how local model-based explanations of neural networks are evaluated. Firstly, they criticize the excessive use of visualization, such as saliency maps, as a means to evaluate explanations\footnote{Other studies have shown that saliency maps are unreliable for evaluating explanations \cite{adebayo2018sanity, geirhos2023don}.}. Moreover, they state that the design principles behind most explanation techniques are not rigorously verified in their respective studies. Thirdly, they criticize the lack of quantifiable measures in some evaluation studies. Lastly, they highlight that while some studies claim to provide explanations that are interpretable to humans, they include limited or no studies that involve human subjects. The study provides general guidelines for improving the quality of research in explaining neural networks. Unlike the study of \cite{leavitt2020towards}, where authors propose general guidelines for a more rigorous study of local model-agnostic explanations, our study aims to showcase the implicit assumptions and the limitations of these evaluation methods used in the literature on eXplainable AI (XAI). 

\section{Local Explanations}
\label{sec:background}
This section briefly defines our formal notion of local model-agnostic explanations\footnote{For brevity, we might refer to local model-agnostic explanations as local explanations in our study.}. Let $X \in \mathbb{R}^{N\times M}$ and $f: \mathbb{R}^{N\times M} \rightarrow \mathbb{R}^{N}$. Let $f(x)$ be the black-box model $f$ predicted output for a designated class given instance $x$, namely $f(x)$. An explanation technique $g$ provides $\phi_j$, the local feature importance of feature $j$ in $x_j$ for the output $f(x)$. The feature importance can be $\phi_j$ zero, which indicates that the feature has no contribution to the predicted output, or negative (positive), which indicates that it negatively (positively) affects the predicted output of black-box of instance $x$, namely $f(x)$. A sub-category of local model-agnostic explanations, e.g. LIME \cite{ribeiro2016should} and SHAP \cite{lundberg2017unified} additionally satisfy the completeness property \cite{lundberg2020local} where

\begin{equation}
    f(x) = \sum_{j=1}^M \phi_j x_j
\label{eq:completneess}
\end{equation}

The completeness property states that the predicted output $f(x)$ equals an additive set of importance scores. The local explanation is created by all the individual feature importance scores into $\Phi_{x,f}^{c} = [\phi_1, ..., \phi_M] \in \mathbb{R}^{M} $. 

Local explanations should not be confused with global explanations. In Local explanations, we explain an individual instance, whereas global explanations provide a single importance scores vector for all instances. In local explanations, each unique instance can have a unique local explanation. On the other hand, global explanations are the feature importance scores for the entire dataset that are equal for all instances. 

\begin{figure}
    \centering
\includegraphics[scale=0.7]{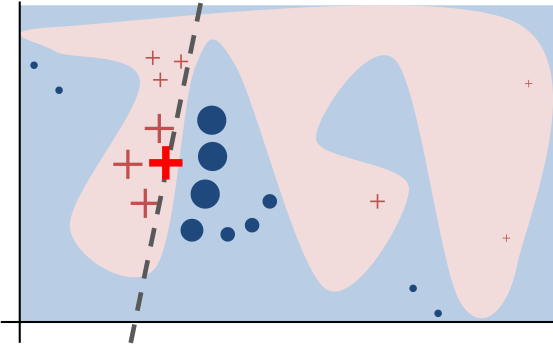}
\caption{LIME explanations are obtained from the weight of the surrogate linear regression model shown by a dotted line. Compared to the original explained model, this model focuses on the model trained in the neighborhood around the explained instance (red bold plus sign). Source: \cite{ribeiro2016should}}
\label{fig:lime_exp_rationale}
\end{figure}

Numerous local model-agnostic explanations, such as LIME \cite{ribeiro2016should}, SHAP \cite{lundberg2017unified} obtain their $\Phi_{x,f}^{c}$  from the weight of an interpretable surrogate $g$. The surrogate model is trained on interpretable representations of explained instances that are interpretable to humans. For example, in the text datasets, binary representations are used where the existence of a token in a sentence is set to one. See Figure \ref{fig:lime_exp_rationale} for an example of how LIME builds its interpretable surrogate.

We want to emphasize that local model-agnostic explanation techniques differ from local (model-based) explanations of neural network models. In those explanations, we need to assume that the model is differentiable since the explanations are obtained based on the derivative of the model $f(x)$ with respect to the input instance $x$, i.e., $df / dx$ \cite{selvaraju2017grad, sundararajan2017axiomatic}. Even though some of the evaluation measures used in evaluating model-based gradient explanations are similar to the ones used for evaluating local model-agnostic explanations, e.g., robustness analysis, we focus only on the taxonomy of the evaluations of local model-agnostic explanations in our study.

\section{History}
\label{sec:history}
The problem of evaluating local explanations is significantly harder than the evaluation of global explanations. This section examines how global explanations have been evaluated in the literature of explainability. Even though these studies have inspired some of the evaluation studies of local explanations, we discuss why these methods cannot be directly used to evaluate local explanations. 

The first approach to evaluate global explanations is by dropping features from the entire dataset and retraining the model \cite{hooker2021unrestricted}. Based on this, the feature importance scores of \textit{accurate} global explanations must equal the difference between the new and original models' accuracy. While this method is imperfect, the evaluation can be considered a controlled experiment. However, this cannot be easily translated into local explanations. Local explanations can be different for each instance. Therefore, dropping an entire set of features is not possible. In \cite{hooker2019benchmark}, the authors propose separately nullifying the important features from each image instance and retraining the model with this new dataset. However, this breaks the properties of a controlled experiment as this process can be sensitive to the type of nullification and the emergence of further complex covariance relationships among the nullified features.

The second approach to evaluating global explanations is based on the fidelity measure. Some studies aimed to replace black-box models with interpretable global explanations, especially in the form of rule classifiers \cite{freitas2014comprehensible, craven1994using, craven1995extracting}. In such studies, the fidelity measure, i.e., the difference between the accuracy of global explanations and the black-box models, showed the quality of those explanations. For the case of evaluating local explanations, the fidelity metric cannot be directly applied. Because in the local explanation techniques, the interpretable surrogate and the black-box models are trained on two different datasets and labels. We want to emphasize that the fidelity measure proposed in \cite{yeh2019fidelity} should not be confused with the measure used in global explanations studies. See section \ref{sec:evaluation:robustness} for more details.

\section{Evaluation Methods of Local Explanations}
\label{sec:evaluation}
In the previous section, we clarified why applying the former approaches to evaluating global explanations is not directly translatable to the problem of evaluating local explanations. As we mentioned in  Section \ref{sec:introduction}, the most straightforward way to evaluate local explanations is to measure their similarity to ground truth importance scores. However, such ground truth needs to be obtained from a black-box model. Remember that we need local explanations because we do not understand black-box models. Therefore, directly evaluating local explanations using local ground truth importance scores is challenging, if not impossible. 

Therefore, all evaluation measures of local explanations need to make certain assumptions. This section provides a taxonomy of evaluation measures, which are categorized based on the assumptions and the ways they circumvent this impossible task. These methods range from Robustness Measures (Section \ref{sec:evaluation:robustness}) and Evaluation based on Ground Truth (Section \ref{sec:evaluation:ground_truth}) to Human-grounded evaluation (Section \ref{sec:evaluation:human}). In each section, we focus on the implicit assumptions and limitations of each measure in a critical manner. In the last section, we provide an example where some evaluation measures from different categories are compared in a synthetic dataset.  

\subsection{Robustness Measures}
\label{sec:evaluation:robustness}
The robustness measures of local explanations can be divided into two sub-categories. In the first category, measures evaluate local explanations by nullifying important (unimportant) features of local explanations. Importance by Preservation, Importance by Deletion \cite{ong2017interpretable} are examples of this first category. The main underlying \textbf{assumption} is that nullifying features deemed important (unimportant) from the local explanation in the explained instance need to cause significant (insignificant) changes in the predicted scores of the explained instance \cite{montavon2018methods}. 

Formally, let $f$ be a black-box model and $E$ the set of top-$K$ features ranked by their importance scores obtained from a local explanation technique $g$ in \textit{descending} order, for instance, $x$. The user selects the variable $K$. Now, Let $x'$ be the explained instance after the features in $E$ are replaced by a baseline value, such as the average feature value in the dataset. The importance By Deletion measure is then measured as $\frac{|f(x') - f(x)|}{|x - x'|}$. Importance by Preservation is calculated similarly; however, in this case, $E$ is the set of top-$K$ features ranked by their importance scores in an \textit{ascending} order. Robust explanations have relatively large (small) Importance by Deletion (Preservation) values \cite{montavon2018methods, alvarez2018robustness}.

In the second category, measures compare the similarity (distance) of the local explanations of the explained instance with the local explanations of an instance that includes a small noise (variation) of the explained instance.  Continuity  \cite{montavon2018methods, agarwal2022rethinking, yeh2019fidelity} is an example of such measures. The second category of the measure is based on the \textbf{assumption} that there needs to be a proportional difference between how much the local explanations of an explained instance changes based on the magnitude of the change in the explained instance.

The continuity measure is an example of this category of robustness measure proposed by \cite{alvarez2018robustness}. Let $x_j$ be an instance located in an Euclidean ball with a maximum radius of $\epsilon$, $B_{\epsilon}(x)$, from the explained instance $x\in X$. We define the continuity for $x_i$ based on the explanation technique $g$ as follows:

\begin{equation}
    \Tilde{L}(x_i) = \underset{x_j \in B_{\epsilon}(x)}{\arg\max}\frac{||g(x) - g(x_j)||_2}{||x - x_j||_2}
\end{equation}

where $\epsilon$ and the size of $B$ is set by the user and $||$ is a norm function.

As it is clear from their definitions, the robustness measures have no further \textbf{assumptions} about the data and model explained.  Moreover, none of these categories of robustness measures include any notion of ground truth for evaluating local explanations. This can make the evaluation process more accessible and can be the reason that they are widely used in evaluating local explanations of black-box models \cite{ribeiro2016should, lundberg2017unified, lundberg2020local, hsieh2021evaluations}. Because of this, we consider them as \textit{indirect} measures for evaluating local explanations.

\begin{figure}
\centering
\includegraphics[scale=0.2]{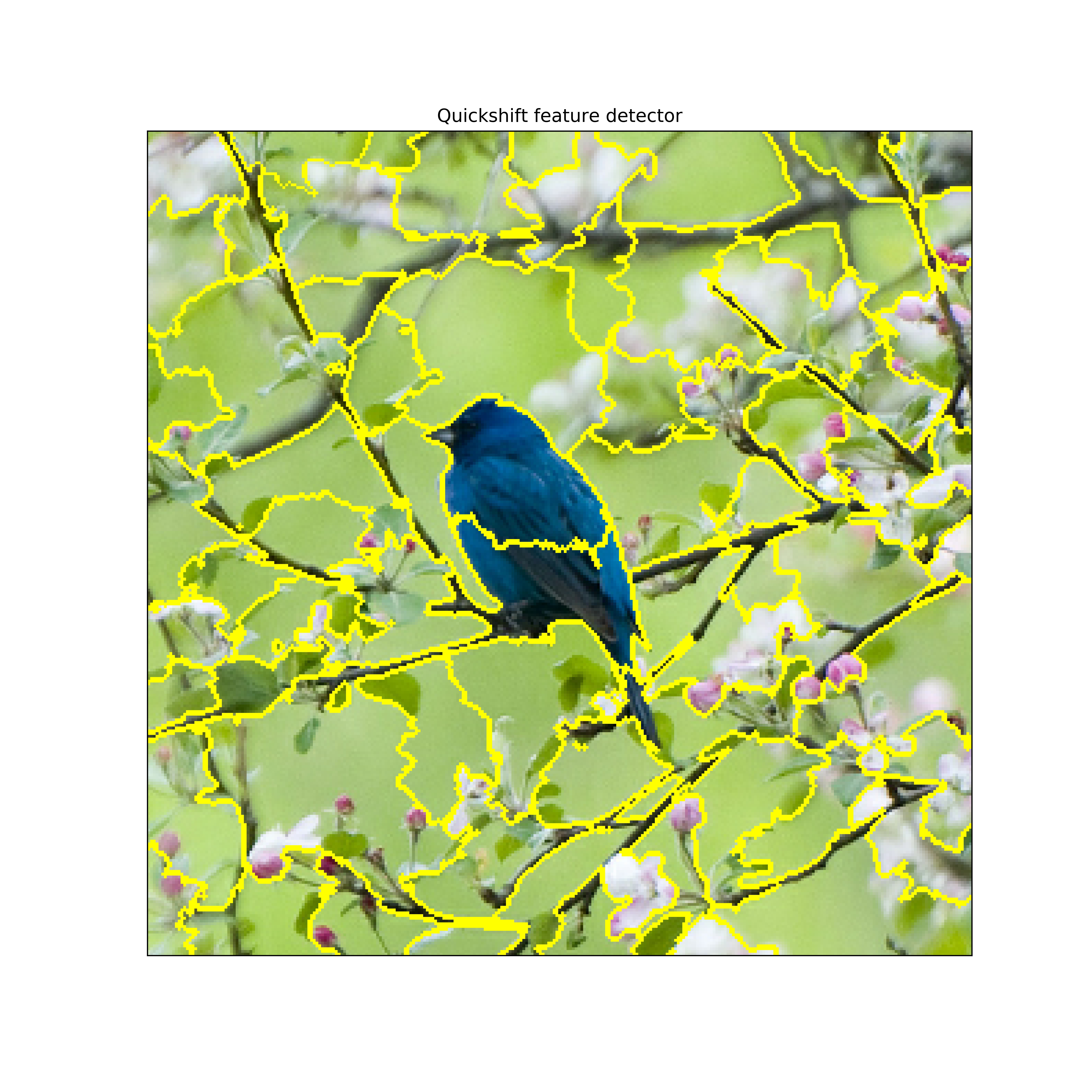}
\includegraphics[scale=0.2]{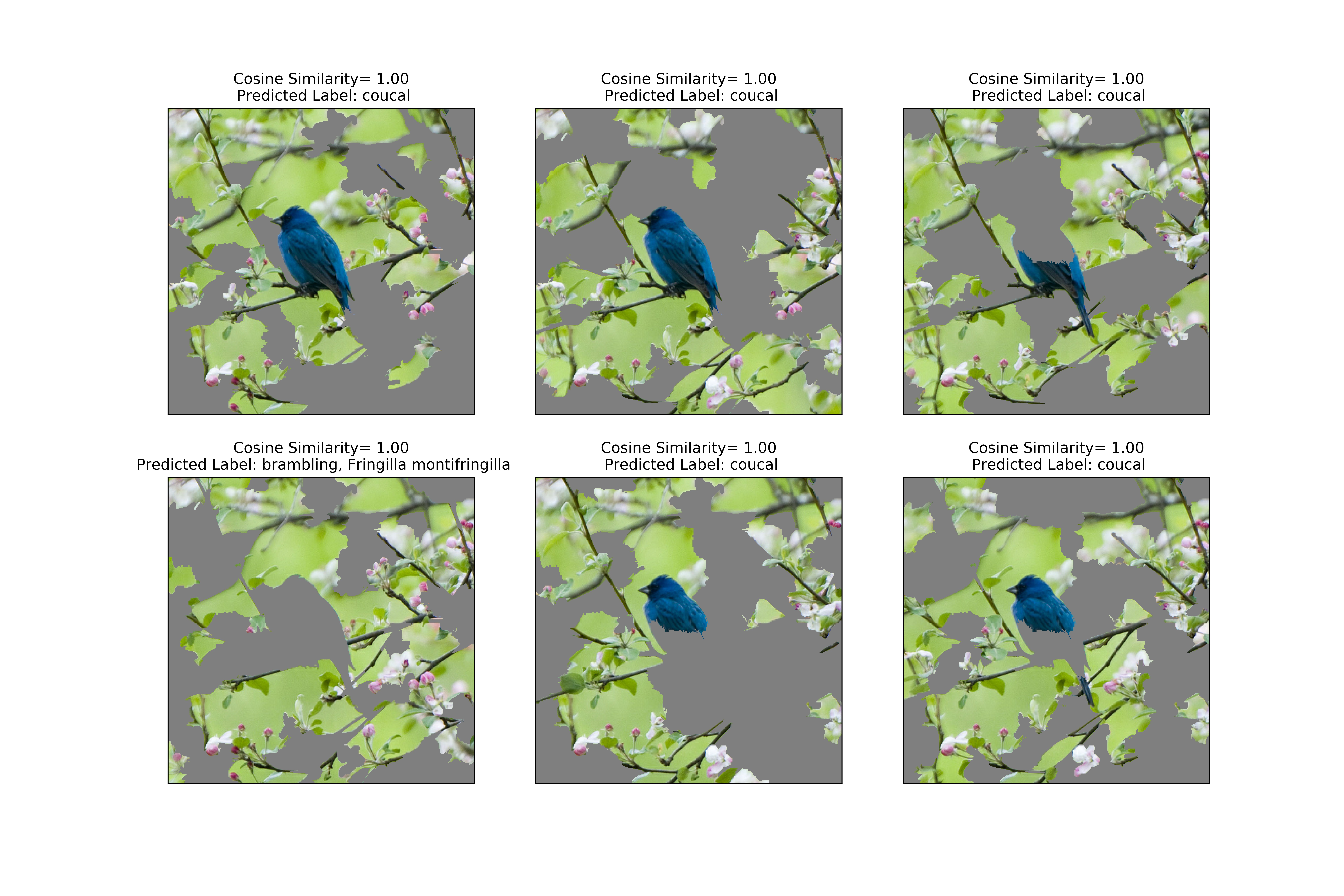}
\caption{(Above): The explained bird image from the ImageNet dataset with the predicted label of \textit{indigo bunting} by the ResNET model. (Below). Let us evaluate the importance of superpixels that include the body of the bird (first image from the bottom row) by removing those pixels. Since the model can still predict the class of the image as a bird, we will wrongfully blame the explanations for inaccurate explanations because of an inaccurate oracle.  Source: \cite{rahnama2019study}}
\label{fig:lime_failure}
\end{figure}

However, they rely heavily on the role of the black-box model as an oracle to provide accurate and certain predictions. Even though we are unaware of studies that have addressed the limitations of the robustness analysis, except partially the work of \cite{strumbelj2010efficient}, we have identified more limitations associated with them:

Firstly and most importantly, \textit{blaming} the local explanation for their lack of robustness is not straightforward. It is equally probable that after nullifying features, the black-box model is providing us with wrongful predictions with high certainty, similar to the case of adversarial examples. We provide examples of this problem in each category of robustness measures. For showing this limitation in the first subcategory of robustness measures, Figure \ref{fig:lime_failure} shows the evaluation process of superpixels, similar to those used by LIME and SHAP explanations. The example includes the image of the class bird with predicted label \textit{indigo bunting}. We can see that nullifying features from these superpixels can generate wrong predictions. For example, in the first image of the bottom row, consider when LIME correctly allocates significant importance scores to the superpixels of the body of the bird. Using Importance by Deletion, we nullify these pixels and record the change in the predicted output of the ResNET model. The model still predicts the label of the instances of the class "bird." Therefore, we blame LIME for inaccurate explanations. We can see that using wrongful information to evaluate explanations may result in blaming explanations for producing explanations that are not robust. However, the model should be blamed for its role as an inaccurate oracle.

\begin{figure}
\centering
\includegraphics[scale=0.4]{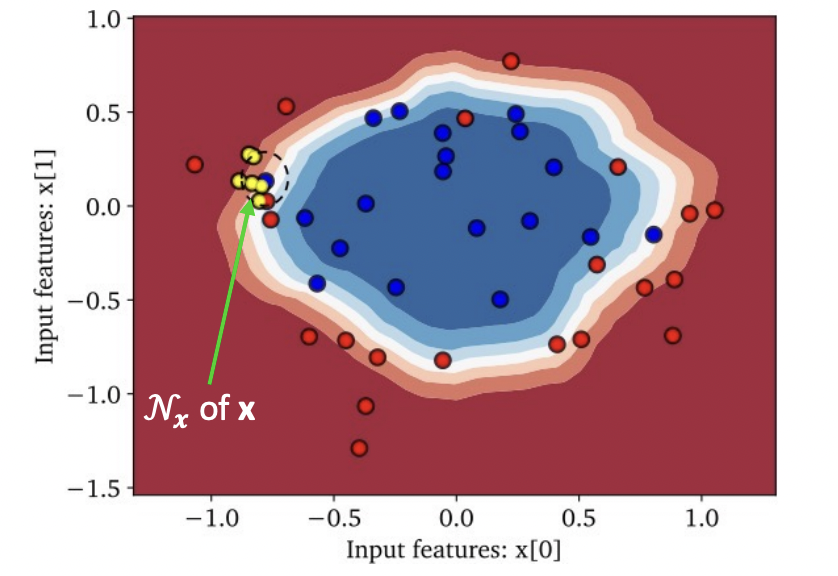}
\caption{For evaluating the robustness of the explained instance (shown with a dotted circle), several instances with added Gaussian noise are created around that instance. Since the explained instance lies close to the decision boundary, it can have a potentially large value for Continuity (hence low robustness concerning this measure). However, the model does not satisfy the Lipshitz condition in the neighborhood around the explained instance. Yet again, we blame explanations for their lack of robustness, whereas the model is at fault. Source: \cite{rahnama2019study}}
\label{fig:continuity_failure}
\end{figure}

To see that the blame problem also exists with the second sub-category of robustness measures, we will show the example provided by \cite{agarwal2022rethinking} (Figure \ref{fig:continuity_failure}). In this example, instances with added Gaussian noise are created to evaluate the Continuity robustness of the explained instance (shown with a dotted circle). Since the instance lies close to the decision boundary, the prediction of these instances will include a significant change in the predicted output and potentially in their local explanations. Because of this, we blame the local explanations for their lack of robustness, yet the underlying reason is that the black-box model does not satisfy the Lipshizt condition around the explained.  

Secondly, there is also a lack of agreement on nullifying features and accounting for that bias in evaluating explanations. In \cite{sturmfels2020visualizing}, the authors show that the choice of the nullification method can severely affect the selection of the most robust explanation technique.

Lastly, there is no global optimal robustness value or an acceptable threshold for selecting robustness explanations. In other words, how much change in the predicted score of a black-box model after nullifying an important (unimportant) feature can deem the explanation robust? \cite{alvarez2018robustness}. In numerous studies, we see how the scale of change after nullifying features can be extremely large between the same model trained on different datasets or two models trained on a single dataset, as shown in \cite{hsieh2021evaluations}. See Table 2 of the study and the scale of values that explanations show for robustness.

Some studies have aimed to address the limitations of the robustness measures. In \cite{hooker2019benchmark, hooker2021unrestricted}, the authors propose to retrain the model after replacing the important (unimportant) features before evaluating the robustness of explanations. Their proposed method aims to tackle the first aforementioned problem above, namely to minimize the problem of uncertain predictions by the oracle. However, this raises the blame problem again: what if the newly trained model does not represent the original black-box model we aimed to explain?

Overall, we need to emphasize that the conclusions we can draw from evaluating local explanations with robustness measures are very limited due to the absence of ground truth. In the next Section, we provide an overview of the methods that introduce the notion of ground truth to evaluate local explanations directly. 

\subsection{Evaluations using Ground Truth}
\label{sec:evaluation:ground_truth}
As mentioned in the previous section, robustness measures can be applied to evaluate local explanations of any model trained on any dataset. However, these measures only measure the accuracy of explanations indirectly. Unlike the robustness measure, using the ground truth-based evaluation measure, we can measure the accuracy of local explanations directly without using the black-box model as an oracle. 

In the upcoming subsections, we show that evaluation based on the ground truth can be categorized into two subcategories: extracting ground truth from synthetic datasets (Section \ref{sec:evaluation:ground_truth_data}) and extracting ground truth from interpretable models (Section \ref{sec:evaluation:ground_truth_model}). In Section \ref{sec:evaluation:similarity_metric}, we show the important role of the similarity metric when evaluating local explanations with these ground truth-based approaches.

\subsubsection{Ground Truth from Synthetic Data}
\label{sec:evaluation:ground_truth_data}
The studies that aim to obtain ground truth from synthetic data are based on the following \textbf{assumption}: if the model we explain is too complex. The extraction of ground truth from them is challenging. However, we can obtain ground truth from synthetic dataset generators. We create synthetic datasets that include prior importance scores for each feature, and then we train the black-box model on this dataset and obtain local explanations. The similarity between the feature importance from the local explanations and the prior importance scores can measure the accuracy of local explanations in this setting. The main benefit of this approach is that we still evaluate the explanations of a black-box model since there is no limitations on the model, and we can also measure the accuracy of local explanations directly.

In \cite{guidotti2021evaluating}, the authors proposed their method Seneca-RC that generates data from a polynomial function that can include varying operators such as $sin$ or $cos$ in its polynomial terms. After that, a sample dataset is generated based on the chosen polynomial function. Lastly, the algorithm returns the ground truth importance scores for the explained instance $x$ based on the following steps: 1) the closest instance $x*$ to $x$ on the decision boundary of an explained model, $f$, is found, and 2) the derivative of the ground truth polynomial is evaluated at this point and returned as true importance scores for $x$. The main benefit of using SenecaRC is that it has a simple logic based on the derivative for various polynomial-based data generation processes.  We show an example of Seneca-RC in Section \ref{section:synthetic_example}. The same study proposes other methods for obtaining ground truth for rule-based, saliency maps, and text-based explanations.

\begin{figure}%
    \centering
    \subfloat[\centering Orange Skin Dataset from \cite{chen2018learning}]{{\includegraphics[width=5cm]{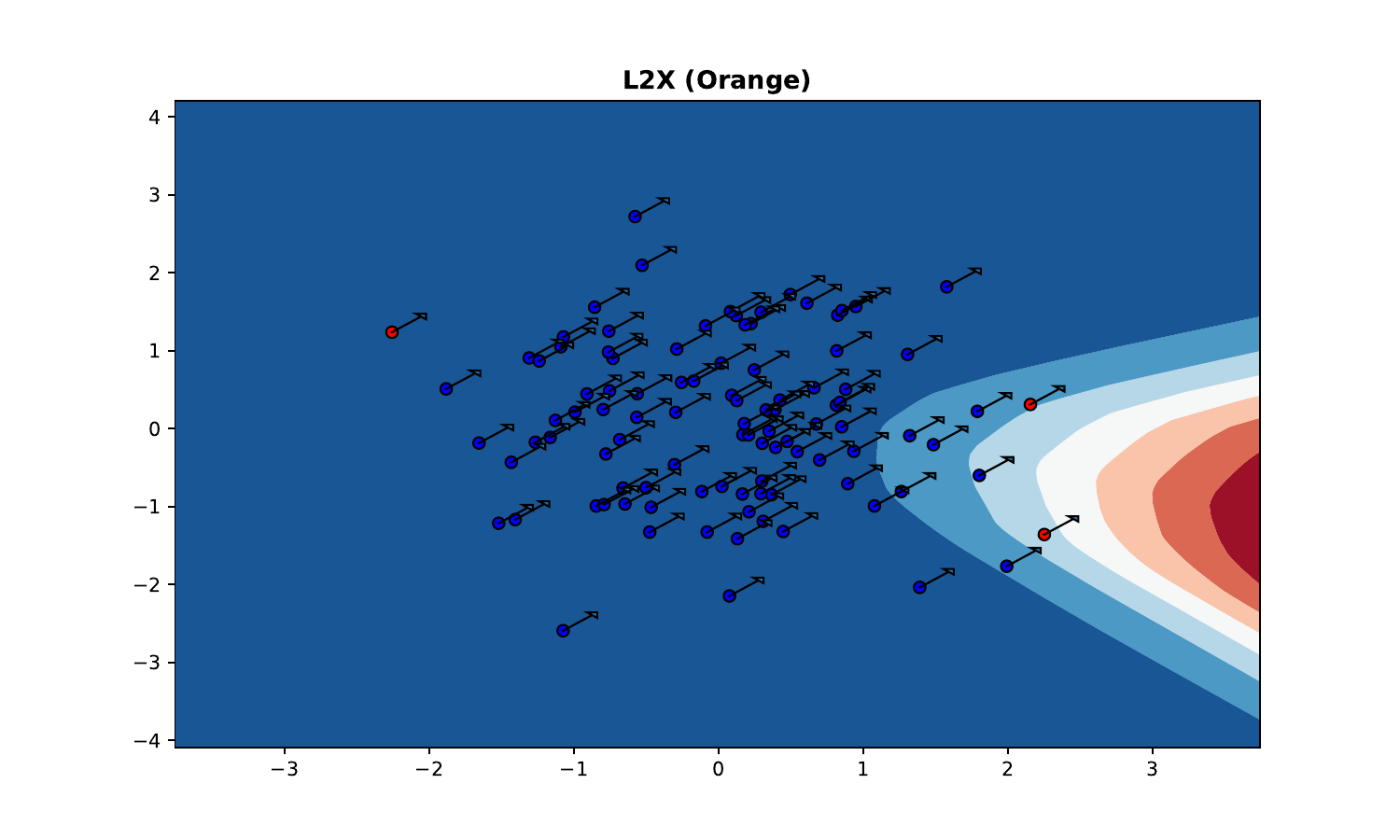} }}%
    \qquad
    \subfloat[\centering Gaussian Dataset from \cite{agarwal2022openxai} (4 clusters)]{{\includegraphics[width=5cm]{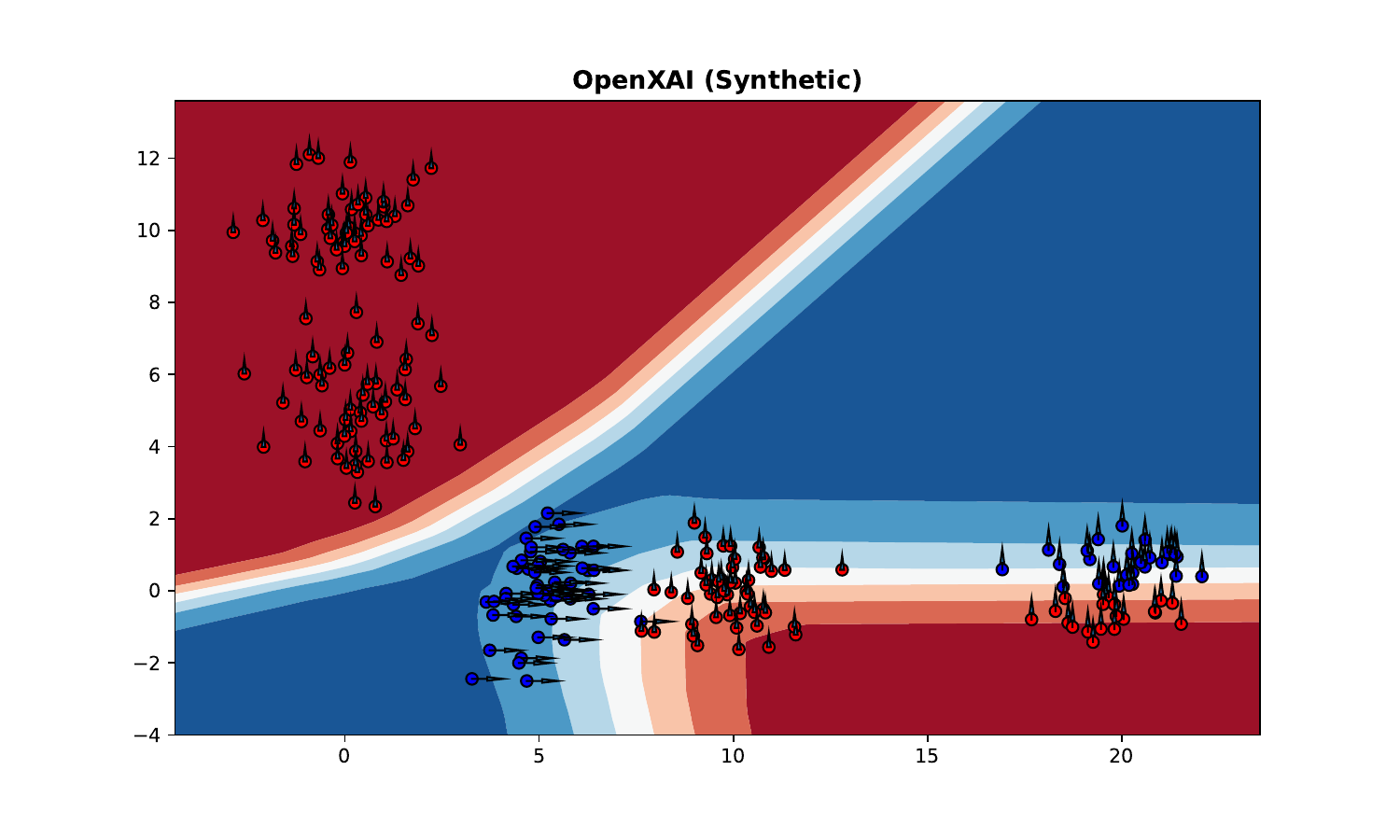} }}%
    \caption{Ground Truth from Synthetic Datasets represented on the decision plane of a multi-layer perception in two trained on this dataset. The arrow represents the ground truth importance scores. Note that the representation of the model does not agree with prior ground truth importance scores.}%
    \label{fig:synthetic_dataset_examples}%
\end{figure}

In \cite{chen2018learning}, the authors evaluate the local explanation techniques using the following synthetic datasets with polynomial features:

\begin{enumerate}
\item $2$-dimensional XOR as binary classification. The input vector $X$ is generated from a $10$-dimensional standard Gaussian. The response variable $Y$ is generated from $P(Y=1|X)\propto \exp \{X_1X_2\}$. 

\item Orange Skin. The input vector $X$ is generated from a $10$-dimensional standard Gaussian. The response variable $Y$ is generated from $P(Y=1|X)\propto \exp \{\sum_{i=1}^4 X_i^2 - 4\}$. Figure \ref{fig:synthetic_dataset_examples} (a) shows an example of this dataset with instance ground truth over the decision plane of Multi-layer perceptron trained on this dataste.

\item Nonlinear additive model. Generate $X$ from a 10-dimensional standard Gaussian. The response variable $Y$ is generated from $P(Y=1|X) \propto \exp\{-100\sin(2X_1)+2|X_2| + X_3 + \exp\{-X_4\}\}$. 

\item Switch feature. Generate $X_{1}$ from a mixture of two Gaussians centered at $\pm 3$ respectively with equal probability. If $X_{1}$ is generated from the Gaussian centered at $3$, the $2-5$th dimensions generate~$Y$ like the orange skin model. Otherwise, the $6-9th$ dimensions are used to generate $Y$ from the nonlinear additive model.   
\end{enumerate}

In this study, after the model is trained on these datasets, local explanations must provide feature importance scores close to the weight of the respective features in the label-generating function. We have found some limitations of the datasets proposed in \cite{chen2018learning}. Firstly, as we can see in Figure \ref{fig:synthetic_dataset_examples}, the ground truth of Orange Skin is equal for all instances irrespective of their position on the decision plane. Similar cases happen in XOR and Nonlinear additive models, as no specific rule changes the label generation formula for specific instances. Even though Switch Feature (dataset 4) can be considered a baseline for local ground truth definitions as the label generation is based on each instance's feature value since the two Gaussian collide around the value of 0, all the features $2-9$ can all be considered important.

In \cite{agarwal2022openxai}, the authors propose a synthetic dataset with Gaussian clusters without covariance between each cluster to solve the problem of datasets such as the Switch feature (Figure \ref{fig:synthetic_dataset_examples} (b)). The dataset allocates random feature masks that control which features can contribute to the predicted output. As shown in the figure, these ground truth importance scores disagree with the model's decision plane. For example, the cluster on the top left disregards the importance of the feature along the x-axis, whereas the nearest decision boundary has used this feature to separate instances, which means that it is important to the model.   

Overall, the main limitation of obtaining ground truth from synthetic datasets in  \cite{guidotti2021evaluating, chen2018learning, agarwal2022openxai} is that it suffers from the blame problem. We are blaming local explanations even though proving that the explained model has learned a representation that follows our prior importance scores is difficult. As highlighted in the work of \cite{chen2020true}, local explanations must be truthful to the model and not the data-generating process. In this case, we again fall into the same problem: we cannot directly blame the explanation technique for inaccurate explanations. 

Another (important) limitation of this category of evaluation measures is that synthetic datasets do not exhibit the complexities of benchmark datasets. Therefore, we do not gain an understanding of the effects complex datasets have on the accuracy of local explanations.   

\subsubsection{Ground Truth from Interpretable Models}
\label{sec:evaluation:ground_truth_model}
One of the limitations of the studies that use ground truth from datasets is that there is no guarantee that the explained model has learned a representation by the data. As mentioned in the previous section, explanation techniques must explain the model, not the data. Some studies have aimed to tackle this limitation by extracting ground truth from models directly, however, from simpler and more interpretable models. Their \textbf{assumption} is as follows: even though we cannot extract ground truth from the black-box model, we can extract them from the more transparent interpretable models. The main benefit of this approach is that we are extracting the ground truth straight from the model's representation and have no assumptions about the datasets. The main strength of this category of evaluation measures is we are confident that these ground truth importance scores are obtained directly from the model and its representation we aim to explain. In these types of evaluations, similar to the methods in Section \ref{sec:evaluation:ground_truth_data}, we can evaluate the local explanations directly.  

In \cite{krishna2022disagreement, agarwal2022openxai}, the ground truth for local explanations is extracted from the weight of Logistic Regression models. Formally,  given weights $w \in \mathbb{R}^{M+1}$  and an instance $x_n \in \mathbb{R}^{M}$, a logistic regression model is defined as:
\begin{equation}
    P(y_n = c || x_n, w) = \frac{1}{1 + e ^{{-\sum_{m=0}^{M} w^m x_n^m}}}
\label{eq:pr_lr}
\end{equation}
\noindent where $x_n^0 = 1$. Based on this, the vector of $w$ is used as the ground truth importance score for all instances. One major drawback of this ground truth is that it is similar for all instances regardless of their feature values.

In \cite{rahnama2023can}, the authors highlight that the approach used by \cite{krishna2022disagreement, agarwal2022openxai} is a baseline for a global explanation and not local explanations as shown earlier in \cite{freitas2014comprehensible, molnar2020interpretable}. Their study proposes extracting the ground truth for local explanation techniques using additive terms of linear additive prediction functions. For example, by transforming the prediction function of Logistic Regression to log odds ratio, they extract the additive terms as the ground truth importance scores. These scores are referred to as Model-Intrinsic Additive Scores (MIAS). More formally, 

\begin{equation}
 log  \frac{P(y_n = c || x_n, w )}{P(y_n = \neg c || x_n, w )} = \sum_{m=0}^{M} w^m x_n^m
 \label{eq:log_lr}
\end{equation}

where $\neg c$ is the complement of class $c$, the authors propose $\lambda_n^m = w^m x_n^m$ as MIAS, the ground truth for local explanations. We can see that in their definition, the feature value of each instance plays a role in the local ground truth importance scores. As we mentioned earlier, local explanations can be unique for different instances, and therefore, the optimal local ground truth needs to include terms specific to each instance as well. The authors show that their proposed method can be used to evaluate local explanations of other interpretable models, such as Linear Regression and Gaussian Naive Bayes. 

There is an advantage to the evaluation methods of this category. Since ground truth importance scores are extracted from the explained model without inducing any change, we can finally blame the inaccuracy of local explanations for themselves. These evaluation methods follow a more principled approach in which they only rely on the explained model, not as an oracle, but as a source to extract ground truth importance scores.

However, there are other limitations associated with this class of evaluation measures. The main limitation of methods that extract ground truth from interpretable models is the explanation techniques were initially developed for explaining black-box models. Therefore, it is not straightforward to conclude that if local explanations accurately explain these interpretable models, they will successfully provide accurate explanations of black-box models. Moreover, ground truth importance scores must be defined separately for each explained model class.  Section \ref{section:synthetic_example} highlights this in an example of a synthetic dataset.

\subsubsection{The role of Similarity Metric}
\label{sec:evaluation:similarity_metric}
Measuring the local explanations directly using ground truth is highly sensitive to the choice of similarity metric. This is a critical issue that has not gained the attention of the studies in the Explainable AI community. The Euclidean distance \cite{alvarez2018robustness}, cosine similarity \cite{ribeiro2016should}, and Spearman's rank correlation 
\cite{ghorbani2019interpretation}, F1-Score \cite{guidotti2021evaluating} are among the set of measures usually used in the evaluation studies of local explanations. 

For illustration, we provide an example of comparing two local explanations using Euclidean and Cosine similarity and Spearman's rank correlation taken from \cite{rahnama2023can}. This example shows that using different similarity metrics can lead to selecting different local explanations based on explanation accuracy. Suppose we need to measure the accuracy of two different local explanations $\phi_1  = [0.21, 0.1, 0.32]$ and  $\phi_2 = [0.21, 0.3, 0.12]$ to the local ground truth score $\lambda= [0.32, 0.2, 0.42]$. We compare the similarity of these explanations with the ground truth:

\noindent\begin{minipage}{.5\linewidth}
\begin{align}
    EuclideanS(\lambda, \phi_1) &= 0.179 \nonumber \\
    SpearmanC(\lambda, \phi_1) &= 1 \nonumber \\ \nonumber 
    CosineS(\lambda, \phi_1) &= 0.99 \nonumber \\ \nonumber 
\end{align}
\end{minipage}%
\begin{minipage}{.5\linewidth}
\begin{align}
    EuclideanS(\lambda, \phi_2) &= 0.28 \nonumber \\
    SpearmanC(\lambda, \phi_2) &= -1 \nonumber  \\
    CosineS(\lambda, \phi_2) &= 0.81\nonumber \\ \nonumber 
\end{align}
\end{minipage}%

Based on Spearman's rank correlation, the ranking of $\phi_1$ correlates perfectly with $\lambda$, while the ranking of $\phi_2$ negatively correlates with $\lambda$. Using this rank-based metric, we can thus conclude that explanation $\phi_1$ is more accurate than $\phi_2$. The Euclidean\footnote{In this example, the Euclidean similarity is defined as $1 / (\epsilon + d)$ where $d$ is the Euclidean distance and $\epsilon$ is the machine epsilon of Python.}, and Cosine Similarity instead votes in favor of $\phi_2$ as the more accurate explanation. This is because Euclidean similarity takes the difference between importance and ground truth scores similarly for all features. On the other hand, cosine similarity only considers the angle between the two vectors. For Spearman's rank correlation, the order of features based on their importance is the most important aspect. 

Even though cosine similarity can be the most optimal measure for data domains such as text and image, we argue that for tabular datasets, rank-based measures such as Spearman's rank correlation might be more suitable for evaluating local explanations. This is large because feature importance is presented to users sorted based on their importance scores in descending order in tabular datasets. In this representation, the similarity among the features with the largest importance scores becomes more important. In addition, the rank-based measures enable comparing feature importance scores with substantially different mechanisms for explanations, e.g., LIME and SHAP versus Permutation Importance. However, the optimal choice of similarity for evaluating local explanations remains an open research question.

\subsection{Evaluation with Model Randomization}
\label{sec:evaluation:model_randomization}
In some studies, local explanations are evaluated based on comparing the local explanations of an accurate black-box model versus after some randomization (contamination) induced on the same black-box model. The \textbf{assumption} is that local explanations need to show significantly different explanations for these two models. In this category of evaluation methods, we no longer have access to or include ground truth in evaluating local explanations.

In \cite{adebayo2018sanity}, the authors propose two randomization tests. In the first test, they randomly re-initialize
the weights of the neural networks model sequentially. In the second test, they independently randomize the weights of a single layer one at a time. They show that most of the local explanation techniques of neural network models, both model-agnostic and model-based, provide similar explanations for the original and randomized models. They conclude that these methods are inaccurate for explaining the investigated neural network models. 

In this category of evaluation measures, we face the blame problem again. Studies have shown that black-box models, including deep neural networks, tend to memorize and extract accurate knowledge even from random or corrupted labels \cite{zhang2021understanding, jiang2018mentornet}. Because of this, there is no guarantee that randomizations can largely obfuscate the workings of the black-box models enough to cause changes to the local explanations. We can blame local explanations in these scenarios even though the explained models can still provide meaningful predictions after introducing randomization.

\subsection{Human-grounded Evaluation}
\label{sec:evaluation:human}
The main focus of our study so far was on the limitations of the functionally grounded evaluations of local explanations. In this section, we briefly describe a set of limitations in the studies that perform human-grounded evaluation of local explanations. 

The human-based evaluations were initially suggested by \cite{kim2016examples}. The authors proposed several ways in which human users can evaluate the explanation. One of the most common methods is called model replay, i.e. a task in which the human subjects are asked to replay the model, i.e., to predict the prediction of the instance explained \cite{poursabzi2021manipulating, ribeiro2016should, nguyen2018comparing} using local explanations. The \textbf{assumption} is that if the human subjects can replay the model promptly, the explanations can be considered accurate. In these studies, it is customary to divide human subjects into experts of the task at hand or lay humans.

One of the main benefits of such methods is that there is also no need to obtain the ground truth for explanations before our evaluation process.On the other hand, there is a limitation associated with them: we cannot subjectively measure how much of the mimicking is performed using human subjects' prior knowledge of data or the model. If human subjects have a poor understanding of the model or data in the task, the poor model replay will still be blamed on local explanations for their inaccuracy. 

In other studies \cite{lundberg2020local}, the notion of consistency with human subjects is considered a metric for evaluating explanations. The measure represents the similarity between human explanations and algorithmic local explanations. The main assumption behind such methods is taht if the similarity is large between the local explanations and human explanations, the local explanations are accurate. However, these methods also suffer from limitations. There are no analysis presented to evaluate whether there are compelete agreements between the logic of the explained model and human subjects in the way they solve the task at hand. Studies have shown that humans and machine learning models rely on different knowledge in performing tasks \cite{arnold2017value}. Because of this, we will blame local explanations again even though the main underlying problem is that the model has learned the task with a significantly different logic. 

\subsection{Synthetic Example}
\label{section:synthetic_example}
As we said earlier, all evaluation measures have different assumptions and study different characteristics of local explanations. In simple words, they are orthogonal to one another. However, in this section, we show an example, taken from \cite{rahnama2023can}, where all of these measures can be compared against one another from synthetic datasets proposed by \cite{guidotti2021evaluating}, the ground truth proposed \cite{agarwal2022openxai}, robustness measures \cite{hsieh2021evaluations, fong2017interpretable} and the MIAS scores of \cite{rahnama2023can}. This is because we use synthetic datasets trained on a Logistic Regression model. 

Let $Y = 2 x_0 - x_1$ be the data generation process where features $x_0$ contribute positively and $x_1$ negatively to the label (Figure \ref{section:synthetic_example}). Let Seneca-RC use this function to generate its synthetic datasets. We sample one thousand instances from Seneca-RC's data generation process where no extra redundant features are added, and we set the noise level to 0.3. We train a Logistic Regression model on this generated dataset. The decision boundary shows that overall, the model has correctly identified that both features are important for separating instances from different classes (Class 1 is represented by the blue color). We also see arrows on top of each instance. The arrows represent the baseline importance scores that each evaluation method uses for evaluating the explanation of local explanations. Note that these arrows do not represent the local explanations but the baselines each evaluation measure uses for evaluating the local explanations.

The Seneca-RC ground truth importance scores are all equal to the vector, $[1, -1]$, irrespective of the position of the instance in the prediction space or the decision boundary of the model. This is because the derivative of the data generation process concerning each feature is a constant value. As mentioned earlier, there can be a discrepancy between the label generation function and the model representation, which is evident in this case.    

The ground truth of OpenXAI \cite{agarwal2022openxai} is also constant across all instances. This is because the model weights are used as the baseline for evaluating all local explanations in this approach. Based on this, all instances receive equal ground truth scores regardless of their position in the decision space. 

Unlike the other methods in our example, robustness measures do not technically have the ground truth for each instance. However, the rationale behind these measures is to measure the effect of nullification of each feature on the prediction of the model's predicted output. Because of this, the arrow on top of instances is created as follows: each feature is nullified separately, and the absolute change in the predicted scores of the model concerning class one is recorded. We have nullified each feature using the average values of that feature in the dataset as it is generally practiced in tabular datasets \cite{liu2021synthetic, montavon2018methods, molnar2022general}.  In the figure, we can see that for most instances, the robustness arrow does not set any importance to the second feature on the y-axis, even though it plays an important role in the linear boundary of logistic regression and the data generation process. Moreover, an instance will receive zero robustness by default along an axis, i.e., for a feature, if its feature values are similar to the empirical average of each feature. This is because nullifying those features will not affect the predicted output.

The Model-Intrinsic Additive Scores (MIAS) allocate different values for instances based on the feature value and their location on the decision plane of the Logistic Regression model. MIAS score of Logistic Regression models sets importance to both features in explaining the log odds ratio of the model. We can also see that the instances will have arrows toward the subspace with maximum log odds of their predicted class visualized by the shades in the background. Moreover, the MIAS vectors of instances close to the decision boundary are more different since the uncertainty in the model's predicted output is larger in those parts of the plane. This satisfied the uniqueness property as discussed in Section \ref{sec:introduction}.

Overall, among the evaluation methods we discussed here, we argue that MIAS scores are the most reliable ground truth for evaluating local explanations of Logistic Regression since 1) Their baseline is sensitive to the decision boundary of the model, and 2) MIAS scores allocate unique ground truth scores for different instances depending on their position in the decision plane of Logistic Regression. 

\begin{figure}[t]
 \centering
  \includegraphics[scale=0.5]{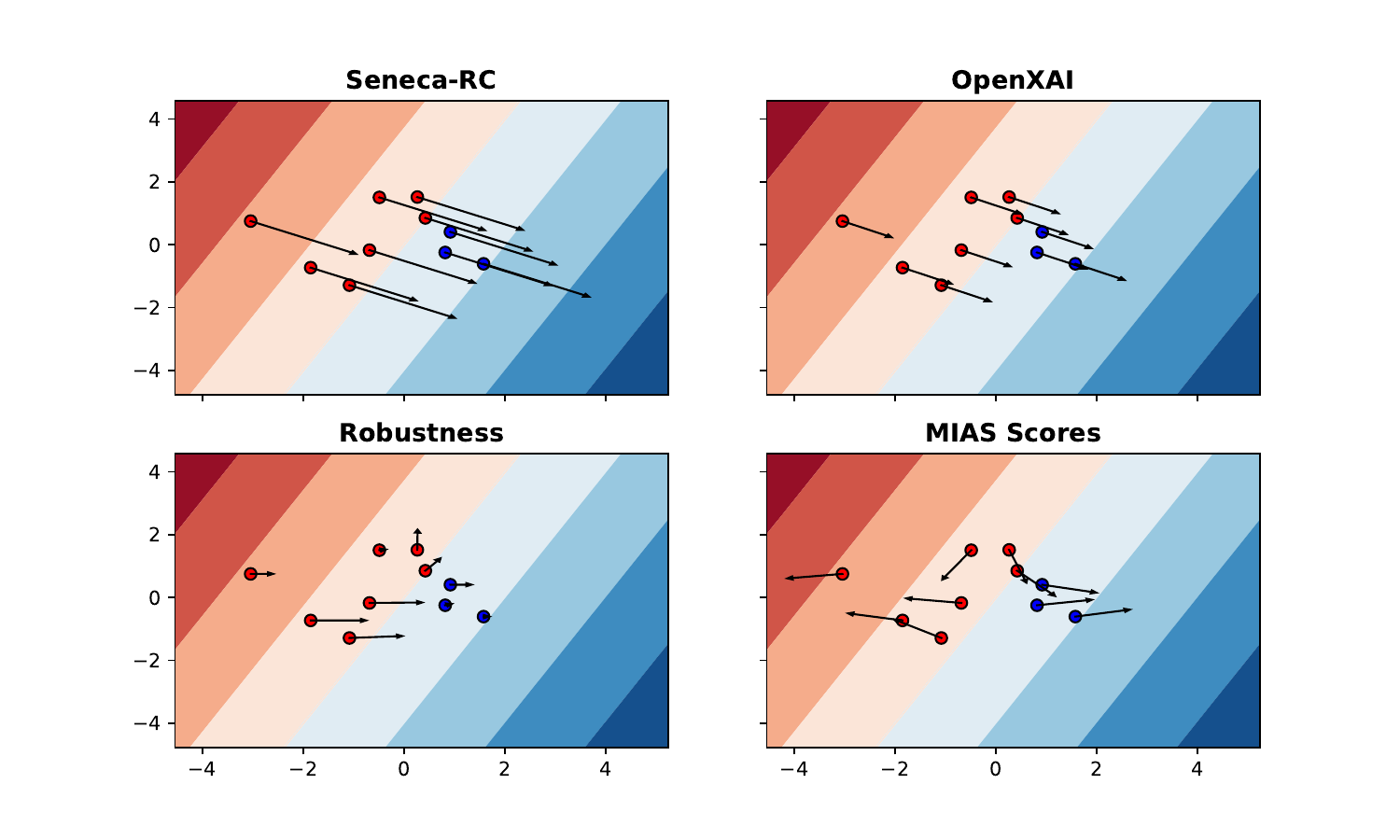}
 \caption{Comparison of the baseline importance scores that Seneca-RC, OpenXAI along with robustness measures and Model-intrinsic Additive Score (MIAS). The Seneca-RC algorithm generated the dataset. The baseline importance score for each instance is visualized as vectors on the top of each instance.}
  \label{fig:synthetic_example_1}
 \end{figure}
 
\section{Conclusion}
\label{sec:conclusion}
Our study presented a taxonomy of the evaluation methods for local model-agnostic explanations: robustness analysis, extracting ground truth from synthetic data and interpretable models, evaluation with model randomization, and human-grounded evaluation. We provide a detailed discussion of each evaluation method's assumptions, strengths, and limitations. Through our study, we highlighted that the significant limitation of all of the categories of evaluation measures is the presence of a "blame problem" where we are unsure of blaming the inaccuracy of local explanations on the explanation techniques or the black-box model or human subjects (in the case of human-based evaluation methods). 

Even though robustness analysis poses no assumption or the type of model or data used for obtaining local explanations, it can only evaluate the local explanations indirectly. The main limitation of robustness analysis is that we can no longer directly blame the explanations for their lack of robustness or the explained model if it provides wrongful predictions. 

Extracting ground truth from a synthetic dataset measures the local explanations directly by setting prior importance scores in the data generation process. However, there are no guarantees that the explained model has learned these prior importance scores. In this case, we will blame the local explanations for inaccuracy instead of the black-box model.

Extracting ground truth from an interpretable model solves the limitation of the synthetic datasets approaches by obtaining ground truth directly from the interpretable model. The number of studies that use this approach is limited, but to our knowledge, they represent the most principled approach to evaluating local model-agnostic explanations. They are the only category of evaluation methods that bypass the blame problem. However, they come with their limitations as well. Since explanation techniques are built to explain black-box models, it is hard to conclude that they are accurate for these models by only looking at local explanations' accuracy of interpretable models.

Evaluation using model randomization assumes that local explanations must provide significantly different explanations after randomizing the model weights or parts of the black-box model. Several studies have shown that randomization in black-box models does not necessarily reduce or change their predictive power. In this case, we can blame the local explanations for their inaccuracy, whereas we need to blame the black-box model.

Using human-grounded evaluation measures to evaluate local explanations can circumvent the need for ground truth importance scores. However, we can end up blaming local explanations for the inherently wrong intuition of the human subjects of the datasets and task at hand or that humans and models are solving the task at hand with different logic. 

The thesis of our study is that none of the available evaluation methods in the literature of explainability is optimal, even though they can circumvent the blame problem in the case of ground truth via interpretable models. Moreover, beyond the blame problem, we need to be aware of the trade-offs these evaluation measures offer. Finding the optimal measure for evaluating local model-agnostic explanations remains an open research problem.  

\bibliography{references}

\begin{thebibliography}{47}
\providecommand{\natexlab}[1]{#1}
\providecommand{\url}[1]{\texttt{#1}}
\expandafter\ifx\csname urlstyle\endcsname\relax
  \providecommand{\doi}[1]{doi: #1}\else
  \providecommand{\doi}{doi: \begingroup \urlstyle{rm}\Url}\fi

\bibitem[Adebayo et~al.(2018)Adebayo, Gilmer, Muelly, Goodfellow, Hardt, and
  Kim]{adebayo2018sanity}
Julius Adebayo, Justin Gilmer, Michael Muelly, Ian Goodfellow, Moritz Hardt,
  and Been Kim.
\newblock Sanity checks for saliency maps.
\newblock \emph{Advances in neural information processing systems}, 31, 2018.

\bibitem[Agarwal et~al.(2022{\natexlab{a}})Agarwal, Johnson, Pawelczyk,
  Krishna, Saxena, Zitnik, and Lakkaraju]{agarwal2022rethinking}
Chirag Agarwal, Nari Johnson, Martin Pawelczyk, Satyapriya Krishna, Eshika
  Saxena, Marinka Zitnik, and Himabindu Lakkaraju.
\newblock Rethinking stability for attribution-based explanations.
\newblock \emph{arXiv preprint arXiv:2203.06877}, 2022{\natexlab{a}}.

\bibitem[Agarwal et~al.(2022{\natexlab{b}})Agarwal, Krishna, Saxena, Pawelczyk,
  Johnson, Puri, Zitnik, and Lakkaraju]{agarwal2022openxai}
Chirag Agarwal, Satyapriya Krishna, Eshika Saxena, Martin Pawelczyk, Nari
  Johnson, Isha Puri, Marinka Zitnik, and Himabindu Lakkaraju.
\newblock Openxai: Towards a transparent evaluation of model explanations.
\newblock \emph{Advances in Neural Information Processing Systems},
  35:\penalty0 15784--15799, 2022{\natexlab{b}}.

\bibitem[Alvarez-Melis and Jaakkola(2018)]{alvarez2018robustness}
David Alvarez-Melis and Tommi~S Jaakkola.
\newblock On the robustness of interpretability methods.
\newblock \emph{arXiv preprint arXiv:1806.08049}, 2018.

\bibitem[Arnold and Kasenberg(2017)]{arnold2017value}
Thomas Arnold and Daniel Kasenberg.
\newblock Value alignment or misalignment {\^a}€“what will keep systems
  accountable?
\newblock In \emph{AAAI Workshop on AI, Ethics, and Society}, 2017.

\bibitem[Chen et~al.(2020)Chen, Janizek, Lundberg, and Lee]{chen2020true}
Hugh Chen, Joseph~D Janizek, Scott Lundberg, and Su-In Lee.
\newblock True to the model or true to the data?
\newblock \emph{arXiv preprint arXiv:2006.16234}, 2020.

\bibitem[Chen et~al.(2018)Chen, Song, Wainwright, and Jordan]{chen2018learning}
Jianbo Chen, Le~Song, Martin Wainwright, and Michael Jordan.
\newblock Learning to explain: An information-theoretic perspective on model
  interpretation.
\newblock In \emph{International Conference on Machine Learning}, pages
  883--892. PMLR, 2018.

\bibitem[Covert et~al.(2021)Covert, Lundberg, and Lee]{covert2021explaining}
Ian Covert, Scott~M Lundberg, and Su-In Lee.
\newblock Explaining by removing: A unified framework for model explanation.
\newblock \emph{J. Mach. Learn. Res.}, 22:\penalty0 209--1, 2021.

\bibitem[Craven and Shavlik(1995)]{craven1995extracting}
Mark Craven and Jude Shavlik.
\newblock Extracting tree-structured representations of trained networks.
\newblock \emph{Advances in neural information processing systems}, 8, 1995.

\bibitem[Craven and Shavlik(1994)]{craven1994using}
Mark~W Craven and Jude~W Shavlik.
\newblock Using sampling and queries to extract rules from trained neural
  networks.
\newblock In \emph{Machine learning proceedings 1994}, pages 37--45. Elsevier,
  1994.

\bibitem[Doshi-Velez and Kim(2017)]{doshi2017towards}
Finale Doshi-Velez and Been Kim.
\newblock Towards a rigorous science of interpretable machine learning.
\newblock \emph{arXiv preprint arXiv:1702.08608}, 2017.

\bibitem[Fong and Vedaldi(2017)]{fong2017interpretable}
Ruth~C Fong and Andrea Vedaldi.
\newblock Interpretable explanations of black boxes by meaningful perturbation.
\newblock In \emph{Proceedings of the IEEE international conference on computer
  vision}, pages 3429--3437, 2017.

\bibitem[Freitas(2014)]{freitas2014comprehensible}
Alex~A Freitas.
\newblock Comprehensible classification models: a position paper.
\newblock \emph{ACM SIGKDD explorations newsletter}, 15\penalty0 (1):\penalty0
  1--10, 2014.

\bibitem[Geirhos et~al.(2023)Geirhos, Zimmermann, Bilodeau, Brendel, and
  Kim]{geirhos2023don}
Robert Geirhos, Roland~S Zimmermann, Blair Bilodeau, Wieland Brendel, and Been
  Kim.
\newblock Don't trust your eyes: on the (un) reliability of feature
  visualizations, 2023.

\bibitem[Ghorbani et~al.(2019)Ghorbani, Abid, and
  Zou]{ghorbani2019interpretation}
Amirata Ghorbani, Abubakar Abid, and James Zou.
\newblock Interpretation of neural networks is fragile.
\newblock In \emph{Proceedings of the AAAI Conference on Artificial
  Intelligence}, volume~33, pages 3681--3688, 2019.

\bibitem[Guidotti(2021)]{guidotti2021evaluating}
Riccardo Guidotti.
\newblock Evaluating local explanation methods on ground truth.
\newblock \emph{Artificial Intelligence}, 291:\penalty0 103428, 2021.

\bibitem[Guidotti et~al.(2021)Guidotti, Monreale, Pedreschi, and
  Giannotti]{guidotti2021principles}
Riccardo Guidotti, Anna Monreale, Dino Pedreschi, and Fosca Giannotti.
\newblock Principles of explainable artificial intelligence.
\newblock In \emph{Explainable AI within the digital transformation and cyber
  physical systems}, pages 9--31. Springer, 2021.

\bibitem[Hancox-Li(2020)]{hancox2020robustness}
Leif Hancox-Li.
\newblock Robustness in machine learning explanations: Does it matter?
\newblock In \emph{Proceedings of the 2020 conference on fairness,
  accountability, and transparency}, pages 640--647, 2020.

\bibitem[Hedstr{\"o}m et~al.(2023)Hedstr{\"o}m, Weber, Krakowczyk, Bareeva,
  Motzkus, Samek, Lapuschkin, and H{\"o}hne]{hedstrom2023quantus}
Anna Hedstr{\"o}m, Leander Weber, Daniel Krakowczyk, Dilyara Bareeva, Franz
  Motzkus, Wojciech Samek, Sebastian Lapuschkin, and Marina M-C H{\"o}hne.
\newblock Quantus: An explainable ai toolkit for responsible evaluation of
  neural network explanations and beyond.
\newblock \emph{Journal of Machine Learning Research}, 24\penalty0
  (34):\penalty0 1--11, 2023.

\bibitem[Hooker et~al.(2021)Hooker, Mentch, and Zhou]{hooker2021unrestricted}
Giles Hooker, Lucas Mentch, and Siyu Zhou.
\newblock Unrestricted permutation forces extrapolation: variable importance
  requires at least one more model, or there is no free variable importance.
\newblock \emph{Statistics and Computing}, 31\penalty0 (6):\penalty0 1--16,
  2021.

\bibitem[Hooker et~al.(2019)Hooker, Erhan, Kindermans, and
  Kim]{hooker2019benchmark}
Sara Hooker, Dumitru Erhan, Pieter-Jan Kindermans, and Been Kim.
\newblock A benchmark for interpretability methods in deep neural networks.
\newblock \emph{Advances in neural information processing systems}, 32, 2019.

\bibitem[Hsieh et~al.(2021)Hsieh, Yeh, Liu, Ravikumar, Kim, Kumar, and
  Hsieh]{hsieh2021evaluations}
Cheng-Yu Hsieh, Chih-Kuan Yeh, Xuanqing Liu, Pradeep Ravikumar, Seungyeon Kim,
  Sanjiv Kumar, and Cho-Jui Hsieh.
\newblock Evaluations and methods for explanation through robustness analysis.
\newblock 2021.

\bibitem[Jiang et~al.(2018)Jiang, Zhou, Leung, Li, and
  Fei-Fei]{jiang2018mentornet}
Lu~Jiang, Zhengyuan Zhou, Thomas Leung, Li-Jia Li, and Li~Fei-Fei.
\newblock Mentornet: Learning data-driven curriculum for very deep neural
  networks on corrupted labels.
\newblock In \emph{International conference on machine learning}, pages
  2304--2313. PMLR, 2018.

\bibitem[Kim et~al.(2016)Kim, Khanna, and Koyejo]{kim2016examples}
Been Kim, Rajiv Khanna, and Oluwasanmi~O Koyejo.
\newblock Examples are not enough, learn to criticize! criticism for
  interpretability.
\newblock \emph{Advances in neural information processing systems}, 29, 2016.

\bibitem[Krishna et~al.(2022)Krishna, Han, Gu, Pombra, Jabbari, Wu, and
  Lakkaraju]{krishna2022disagreement}
Satyapriya Krishna, Tessa Han, Alex Gu, Javin Pombra, Shahin Jabbari, Steven
  Wu, and Himabindu Lakkaraju.
\newblock The disagreement problem in explainable machine learning: A
  practitioner's perspective.
\newblock \emph{arXiv preprint arXiv:2202.01602}, 2022.

\bibitem[Leavitt and Morcos(2020)]{leavitt2020towards}
Matthew~L Leavitt and Ari Morcos.
\newblock Towards falsifiable interpretability research.
\newblock \emph{arXiv preprint arXiv:2010.12016}, 2020.

\bibitem[Liu et~al.(2021)Liu, Khandagale, White, and
  Neiswanger]{liu2021synthetic}
Yang Liu, Sujay Khandagale, Colin White, and Willie Neiswanger.
\newblock Synthetic benchmarks for scientific research in explainable machine
  learning.
\newblock \emph{arXiv preprint arXiv:2106.12543}, 2021.

\bibitem[Lundberg and Lee(2017)]{lundberg2017unified}
Scott~M Lundberg and Su-In Lee.
\newblock A unified approach to interpreting model predictions.
\newblock \emph{Advances in neural information processing systems}, 30, 2017.

\bibitem[Lundberg et~al.(2020)Lundberg, Erion, Chen, DeGrave, Prutkin, Nair,
  Katz, Himmelfarb, Bansal, and Lee]{lundberg2020local}
Scott~M Lundberg, Gabriel Erion, Hugh Chen, Alex DeGrave, Jordan~M Prutkin,
  Bala Nair, Ronit Katz, Jonathan Himmelfarb, Nisha Bansal, and Su-In Lee.
\newblock From local explanations to global understanding with explainable ai
  for trees.
\newblock \emph{Nature machine intelligence}, 2\penalty0 (1):\penalty0 56--67,
  2020.

\bibitem[Molnar et~al.(2020)Molnar, Casalicchio, and
  Bischl]{molnar2020interpretable}
Christoph Molnar, Giuseppe Casalicchio, and Bernd Bischl.
\newblock Interpretable machine learning--a brief history, state-of-the-art and
  challenges.
\newblock In \emph{Joint European Conference on Machine Learning and Knowledge
  Discovery in Databases}, pages 417--431. Springer, 2020.

\bibitem[Molnar et~al.(2022)Molnar, K{\"o}nig, Herbinger, Freiesleben, Dandl,
  Scholbeck, Casalicchio, Grosse-Wentrup, and Bischl]{molnar2022general}
Christoph Molnar, Gunnar K{\"o}nig, Julia Herbinger, Timo Freiesleben, Susanne
  Dandl, Christian~A Scholbeck, Giuseppe Casalicchio, Moritz Grosse-Wentrup,
  and Bernd Bischl.
\newblock General pitfalls of model-agnostic interpretation methods for machine
  learning models.
\newblock In \emph{International Workshop on Extending Explainable AI Beyond
  Deep Models and Classifiers}, pages 39--68. Springer, 2022.

\bibitem[Montavon et~al.(2018)Montavon, Samek, and
  M{\"u}ller]{montavon2018methods}
Gr{\'e}goire Montavon, Wojciech Samek, and Klaus-Robert M{\"u}ller.
\newblock Methods for interpreting and understanding deep neural networks.
\newblock \emph{Digital Signal Processing}, 73:\penalty0 1--15, 2018.

\bibitem[Nauta et~al.(2023)Nauta, Trienes, Pathak, Nguyen, Peters, Schmitt,
  Schl{\"o}tterer, van Keulen, and Seifert]{nauta2023anecdotal}
Meike Nauta, Jan Trienes, Shreyasi Pathak, Elisa Nguyen, Michelle Peters,
  Yasmin Schmitt, J{\"o}rg Schl{\"o}tterer, Maurice van Keulen, and Christin
  Seifert.
\newblock From anecdotal evidence to quantitative evaluation methods: A
  systematic review on evaluating explainable ai.
\newblock \emph{ACM Computing Surveys}, 55\penalty0 (13s):\penalty0 1--42,
  2023.

\bibitem[Nguyen(2018)]{nguyen2018comparing}
Dong Nguyen.
\newblock Comparing automatic and human evaluation of local explanations for
  text classification.
\newblock In \emph{Proceedings of the 2018 Conference of the North American
  Chapter of the Association for Computational Linguistics: Human Language
  Technologies, Volume 1 (Long Papers)}, pages 1069--1078, 2018.

\bibitem[Poursabzi-Sangdeh et~al.(2021)Poursabzi-Sangdeh, Goldstein, Hofman,
  Wortman~Vaughan, and Wallach]{poursabzi2021manipulating}
Forough Poursabzi-Sangdeh, Daniel~G Goldstein, Jake~M Hofman, Jennifer~Wortman
  Wortman~Vaughan, and Hanna Wallach.
\newblock Manipulating and measuring model interpretability.
\newblock In \emph{Proceedings of the 2021 CHI conference on human factors in
  computing systems}, pages 1--52, 2021.

\bibitem[Rahnama and Bostr{\"o}m(2019)]{rahnama2019study}
Amir Hossein~Akhavan Rahnama and Henrik Bostr{\"o}m.
\newblock A study of data and label shift in the lime framework.
\newblock \emph{arXiv preprint arXiv:1910.14421}, 2019.

\bibitem[Rahnama et~al.(2023)Rahnama, B{\"u}tepage, Geurts, and
  Bostr{\"o}m]{rahnama2023can}
Amir Hossein~Akhavan Rahnama, Judith B{\"u}tepage, Pierre Geurts, and Henrik
  Bostr{\"o}m.
\newblock Can local explanation techniques explain linear additive models?
\newblock \emph{Data Mining and Knowledge Discovery}, pages 1--44, 2023.

\bibitem[Ribeiro et~al.(2016{\natexlab{a}})Ribeiro, Singh, and
  Guestrin]{ribeiro2016model}
Marco~Tulio Ribeiro, Sameer Singh, and Carlos Guestrin.
\newblock Model-agnostic interpretability of machine learning.
\newblock \emph{arXiv preprint arXiv:1606.05386}, 2016{\natexlab{a}}.

\bibitem[Ribeiro et~al.(2016{\natexlab{b}})Ribeiro, Singh, and
  Guestrin]{ribeiro2016should}
Marco~Tulio Ribeiro, Sameer Singh, and Carlos Guestrin.
\newblock " why should i trust you?" explaining the predictions of any
  classifier.
\newblock In \emph{Proceedings of the 22nd ACM SIGKDD international conference
  on knowledge discovery and data mining}, pages 1135--1144,
  2016{\natexlab{b}}.

\bibitem[Rudin(2018)]{rudin2018please}
Cynthia Rudin.
\newblock Please stop explaining black box models for high stakes decisions.
\newblock \emph{Stat}, 1050:\penalty0 26, 2018.

\bibitem[Selvaraju et~al.(2017)Selvaraju, Cogswell, Das, Vedantam, Parikh, and
  Batra]{selvaraju2017grad}
Ramprasaath~R Selvaraju, Michael Cogswell, Abhishek Das, Ramakrishna Vedantam,
  Devi Parikh, and Dhruv Batra.
\newblock Grad-cam: Visual explanations from deep networks via gradient-based
  localization.
\newblock In \emph{Proceedings of the IEEE international conference on computer
  vision}, pages 618--626, 2017.

\bibitem[Strumbelj and Kononenko(2010)]{strumbelj2010efficient}
Erik Strumbelj and Igor Kononenko.
\newblock An efficient explanation of individual classifications using game
  theory.
\newblock \emph{The Journal of Machine Learning Research}, 11:\penalty0 1--18,
  2010.

\bibitem[Sturmfels et~al.(2020)Sturmfels, Lundberg, and
  Lee]{sturmfels2020visualizing}
Pascal Sturmfels, Scott Lundberg, and Su-In Lee.
\newblock Visualizing the impact of feature attribution baselines.
\newblock \emph{Distill}, 5\penalty0 (1):\penalty0 e22, 2020.

\bibitem[Sundararajan et~al.(2017)Sundararajan, Taly, and
  Yan]{sundararajan2017axiomatic}
Mukund Sundararajan, Ankur Taly, and Qiqi Yan.
\newblock Axiomatic attribution for deep networks.
\newblock In \emph{International conference on machine learning}, pages
  3319--3328. PMLR, 2017.

\bibitem[Yeh et~al.(2019)Yeh, Hsieh, Suggala, Inouye, and
  Ravikumar]{yeh2019fidelity}
Chih-Kuan Yeh, Cheng-Yu Hsieh, Arun Suggala, David~I Inouye, and Pradeep~K
  Ravikumar.
\newblock On the (in) fidelity and sensitivity of explanations.
\newblock \emph{Advances in Neural Information Processing Systems}, 32, 2019.

\bibitem[Zhang et~al.(2021)Zhang, Bengio, Hardt, Recht, and
  Vinyals]{zhang2021understanding}
Chiyuan Zhang, Samy Bengio, Moritz Hardt, Benjamin Recht, and Oriol Vinyals.
\newblock Understanding deep learning (still) requires rethinking
  generalization.
\newblock \emph{Communications of the ACM}, 64\penalty0 (3):\penalty0 107--115,
  2021.

\bibitem[Zhou et~al.(2021)Zhou, Gandomi, Chen, and
  Holzinger]{zhou2021evaluating}
Jianlong Zhou, Amir~H Gandomi, Fang Chen, and Andreas Holzinger.
\newblock Evaluating the quality of machine learning explanations: A survey on
  methods and metrics.
\newblock \emph{Electronics}, 10\penalty0 (5):\penalty0 593, 2021.

\end{thebibliography}
\end{document}